# Standards for Language Resources

Nancy Ide,* Laurent Romary[†]

* Department of Computer Science
Vassar College
Poughkeepsie, New York 12604-0520 USA
ide@cs.vassar.edu

[†] Equipe Langue et Dialogue
LORIA/INRIA
Vandoeuvre-lès Nancy, FRANCE
romary@loria.fr

**Abstract**
This paper presents an abstract data model for linguistic annotations and its implementation using XML, RDF and related standards; and to outline the work of a newly formed committee of the International Standards Organization (ISO), ISO/TC 37/SC 4 Language Resource Management, which will use this work as its starting point. The primary motive for presenting the latter is to solicit the participation of members of the research community to contribute to the work of the committee.

## 1. Introduction

The goal of this paper is two-fold: to present an abstract data model for linguistic annotations and its implementation using XML, RDF and related standards; and to outline the work of a newly formed committee of the International Standards Organization (ISO), ISO/TC 37/SC 4 Language Resource Management, which will use this work as its starting point. The primary motive for presenting the latter is to solicit the participation of members of the research community to contribute to the work of the committee.

The objective of ISO/TC 37/SC 4 is to prepare international standards and guidelines for effective language resource management in applications in the multilingual information society. To this end, the committee will develop principles and methods for creating, coding, processing and managing language resources, such as written corpora, lexical corpora, speech corpora, dictionary compiling and classification schemes. The focus of the work is on data modeling, markup, data exchange and the evaluation of language resources other than terminologies (which have already been treated in ISO/TC 37). The worldwide use of ISO/TC 37/SC 4 standards should improve information management within industrial, technical and scientific environments, and increase efficiency in computer-supported language communication.

## 2. Motivation

The standardization of principles and methods for the collection, processing and presentation of language resources requires a distinct type of activity. Basic standards must be produced with wide-ranging applications in view. In the area of language resources, these standards should provide various technical committees of ISO, IEC and other standardizing bodies with the groundwork for building more precise standards for language resource management.

The need for harmonization of representation formats for different kinds of linguistic information is critical, as resources and information are more and more frequently merged, compared, or otherwise utilized in common systems. This is perhaps most obvious for processing multi-modal information, which must support the fusion of multimodal inputs and represent the combined and integrated contributions of different types of input (e.g., a spoken utterance combined with gesture and facial expression), and enable multimodal output (see, for example, Bunt and Romary, 2002). However, language processing applications of any kind require the integration of varieties of linguistic information, which, in today's environment, come from potentially diverse sources. We can therefore expect use and integration of, for example, syntactic, morphological, discourse, etc. information for multiple languages, as well as information structures like domain models and ontologies.

We are aware that standardization is a difficult business, and that many members of the targeted communities are skeptical about imposing any sort of standards at all. There are two major arguments against the idea of standardization for language resources. First, the diversity of theoretical approaches to, in particular, the annotation of various linguistic phenomena suggests that standardization is at least impractical, if not impossible. Second, it is feared that vast amounts of existing data and processing software, which may have taken years of effort and considerable funding to develop, will be rendered obsolete by the acceptance of new standards by the community. To answer both of these concerns, we stress that the efforts of the committee are geared toward defining *abstract models* and *general frameworks* for creation and representation of language resources, rather than specific formats. These models should, in principle, be abstract enough to accommodate diverse theoretical approaches. The model so far developed in ISO TC/37 for terminology, which has informed and been informed by work on representation schemes for dictionaries and other lexical data (Ide, *et al.*, 2000) and syntactic annotation (Ide & Romary, 2001) demonstrates that this is not an unrealizable goal. Also, by situating all of the standards development squarely in the framework of XML and related standards such as RDF, DAML+OIL, etc., we hope to ensure not only that the standards developed by the committee provide for compatibility with established and widely accepted web-based technologies, but also that transduction from legacy formats into XML formats conformant to the new standards is feasible.

ISO/TC 37/SC 4 will liaison with ISLE (International Standards for Language Engineering), which has implemented various recent efforts to integrate EC and US efforts for language resources. Where possible, these and other standards set up in EAGLES will be incorporated into the ISO standards. ISO/TC 37/SC 4 will also broaden the work of EAGLES/ISLE by including languages (e.g. Asian languages) that are not currently covered by EAGLES/ISLE standards.

At present, language professionals and standardization experts are not sufficiently aware of the standardization efforts being undertaken by ISO/TC 37/SC 4. Promoting awareness of future activities and rising problems, therefore, will be a crucial factor in the success of the committee, and will be required to ensure widespread adoption of the standards it develops. An even more critical factor for the success of the committee's work is to involve, from the outset, as many and as broad a range of potential users of the standards as possible. This presentation serves as a call for participation to the linguistics and computational linguistics research communities.

## 3. Objectives

ISO TC37/SC 4's goal is to develop a platform for the design and implementation of linguistic resource formats and processes in order to facilitate the exchange of information between language processing modules. This will be accomplished by defining a *common interface format* capable of representing multiple kinds of linguistic information. The interface format must support the communication among all modules in the system, and be adequate for representing not only the end result of interpretation, but also intermediate results.

A well-defined representational framework for linguistic information will also provide for the specification and comparison of existing application-specific representations and the definition of new ones, while ensuring a level of interoperability between them.

### 3.1. Requirements

Very generally, a linguistic representation framework must meet the following requirements:

*Expressive adequacy:* the framework should be expressive enough to represent all varieties of linguistic information;

*Semantic adequacy:* the representation structures should have a formal semantics, i.e., their definition should provide a rigorous basis for further processing (e.g., deductive reasoning, statistical analysis, generation, etc.).

Providing interface formats within a system architecture demands that "incremental" construction of intermediate and partial representations be supported. In addition, if the construction of a final representation does not succeed, the representation must capture the information required to enable appropriate system action. This dictates additional requirements:

*Incrementality:* support for the various stages of input interpretation and output generation, allowing both early and late fusion and fission.

*Uniformity:* the representation of various types of input and output should utilize the same "building blocks" and the same methods for combining complex structures composed of these building blocks.

*Underspecification and partiality*: support for the representation of partial and intermediate results, including the capture of unresolved ambiguities.

Finally, the representational framework must be accommodate the developing field of language processing system design by satisfying these further requirements:

*Openness:* the framework should not depend on a single linguistic theory, but should enable representations based on different theories and approaches;

*Extensibilty*. The framework should be compatible with alternative methods for designing representation schemas (e.g., XML) rather than being tied to a specific schema.

### 3.2. Methodology

A working group of SC 4 (WG1/WI-1) has been charged with the task of defining a linguistic annotation framework, which will be used by other SC 4 working groups to develop more precise specifications for particular annotation types. The full list of SC 4 working groups is as follows:

WG1/WI-0: Terminology for Language Resources
WG1/WI-1: Linguistic annotation framework
WG1/WI-2: Meta-data for multimodal and multilingual information
WG2/WI-3: Structural content representation (syntax and morphology)
WG2/WI-4: Multimodal content representation
WG2/WI-5: Discourse level representation
WG3/WI-6a: Multilingual text representation
WG4/WI-7: Lexicons
WG5/WI-8: Validation of language resources
WG5/WI-9: Net-based distributed cooperative work for the creation of LRs

We focus here on the work of WG1/WI-1, which will serve as the starting point for that of most of the others. This group will propose a *data architecture* consisting of basic mechanisms and data structures for linguistic annotation and representation, comprised of the following:

*Basic components*: the basic constructs for building representations of linguistic information; specifically, identification of types of building blocks and ways to connect them.

*General mechanisms*: representation techniques that make the annotations more compact and flexible and enable linking them to external sources of information; for example, sub-structure labeling, argument under-specification, restrictions on label values and/or disjunctions or lists to represent ambiguity or partiality, structure sharing; linking to domain models, linking to other levels of annotation, etc.

*Contextual data categories*: administrative (meta-) data relevant for processing, such as environment data (e.g., time stamps, spatial information); processing information (e.g., module that produced the representation; confidence level); interaction information (speaker, audience, etc.).

The following section outlines a linguistic framework which will serve as the starting point for development within SC 4. The current model is based on work on development of annotation formats for lexicons (Ide, et al., 2001), morphosyntactic and syntactic annotation (Ide & Romary, 2001a; Ide & Romary, 2001b; Ide & Romary, forthcoming), and which has been further developed within TC37/SC4 for the definition of TMF (Terminological Markup Framework; ISO 16642, under DIS ballot).

## 4. A Framework for Linguistic Annotation

Our fundamental assumption is that representation formats for linguistic data and its annotations can be modeled by combining a structural *meta-model,* that is, an abstract structure shared by all documents of a given type (e.g. syntactic annotation), with a set of *data categories* that are associated with the various components of the meta-model. Our work in SC4 is concerned, first, with identification of a reduced set of meta-models that can be used for any type of linguistic data and its annotations. Data categories, on the other hand, are defined by the implementer; interoperability among formats is ensured by providing a *Data Category Registry* in which the categories and relations required for a particular type of annotation are precisely defined.

The model for linguistic annotation must satisfy two general criteria:

1. It must be possible to instantiate it using a standard representational format;
2. It must be designed so as to serve as a pivot format into and out of which proprietary formats can be transduced, in order to enable comparison and merging, as well as operation on the data by common tools.

### 4.1. Abstract model for annotation

At its highest level of abstraction, an annotation is a set of data or information (in our case, linguistic information) that is associated with some other data. The latter is what could be called "primary" data (e.g., a part of a text or speech signal, etc.), but this need not be the case; consider, for example, the alignment of parallel translations, where the "annotation" is a link between two primary data objects (the aligned texts). Typically, primary data objects are represented by "locations" in an electronic file, for example, the span of characters comprising a sentence or word, or a point at which a given temporal event begins or ends (as in speech annotation). As such, at the base primary data objects are relatively simple in their structure; more complex data objects may consist of a list or set of contiguous or non-contiguous locations. Annotation objects, on the other hand, often have a more complex internal structure: syntactic annotation, for example, may be expressed as a tree structure, and may include more elemental annotations such as dependency relations (which is itself an annotation relating two objects, where the relation is directional (dependent-to-head)).

Thus, we can conceive of an annotation as a one- or two-way link between an annotation object and a point (or a list/set of points) or span (or a list/set of spans) within a base data set. Links may or may not have a semantics-- i.e., a type--associated with them. Points and spans in the base data may themselves be objects, or sets or lists of objects. This abstract formulation can serve as the basis for defining a general model for linguistic annotation that can be realized in a standard representational format. In fact, this model is consistent with well-established data modeling concepts used in diverse areas, including knowledge representation (KR), object-oriented design, and database systems, and which inform fundamental data structures in computer science (trees, graphs, etc.) and database design (notably, the Entity-Relationship (ER) model). As such, the model provides us with established means to describe our data objects (in terms of composition, attributes, class membership, applicable procedures, etc.) and relations among them, independent of their instantiation in any particular form. It also ensures that standardized representation formats exist that can instantiate the model.

One way to represent linguistic annotation in terms of the abstract model is as a graph of elementary *structural nodes* to which one or more *information units* are attached. The distinction between the structure of annotations and the informational units of which it is comprised is, we feel, critical to the design of a truly general model for annotations. Annotations may be structured in several ways; perhaps the most common structure is hierarchical. For example, phrase structure analyses of syntax are structured as trees; in addition, hierarchy is often used to break annotation information into sub-components, as in the case of lexical and terminological information.

There are several special relations *among* annotations that must be represented in the model, including the following:

*Parallelism:* two or more annotations refer to the same data object;

*Alternatives:* two or more annotations comprise a set of mutually exclusive alternatives (e.g., two possible part-of-speech assignments, before disambiguation);

*Aggregation:* two or more annotations comprise a list or set that should be taken as a unit.

Information units or *data categories* provide the semantics of the annotation. Data categories are the most theory and application-specific part of an annotation scheme. We do not attempt to define the relevant data categories for given types of annotation. Rather, we propose the development of a Data Category Registry to provide a framework in which the research community can formally define data categories for reference and use in annotation. To make them maximally interoperable and consistent with existing standards, data categories can be defined using RDF schemas to formalize the properties and relations associated with each. Note that RDF descriptions function much like class definitions in an object-oriented programming language: they provide, effectively, templates that describe how objects may be instantiated, but do not constitute the objects themselves. Thus, in a document containing an actual annotation, several objects with the same type may be instantiated, each with a different value. The RDF schema ensures that each instantiation is recognized as a sub-class of more general classes and inherits the appropriate properties.

A formally defined set of categories will have several functions: (1) it will provide a precise semantics for annotation categories that can be either used "off the shelf" by annotators or modified to serve specific needs; (2) it will provide a set of reference categories onto which scheme-specific names can be mapped; and (3) it will provide a point of departure for definition of variant or more precise categories. Thus the overall goal of the Data Category Registry is not to impose a specific set of categories, but rather to ensure that the semantics of data categories included in annotations (whether they exist in the Registry or not) are well-defined and understood.

## 5. An Example

We illustrate a simple application of the framework presented above for the domain of morpho-syntactic annotation. For the purposes of illustration, it is necessary to make technical choices concerning the representation format. XML and related standards developed by the World Wide Web consortium appear at present to provide the best means to represent information structures intended to be transmitted across a network. For the purposes of linguistic resource representation, XML provides several important features:

- it is both Unicode and ISO 10646 compatible;
- XML namespaces provide the options of combining element definitions from multiple sources in an XML document, thereby fostering modularity and reuse;
- XML schemas provide a powerful means to define, constrain, and extend definitions of the structure and contents of classes of XML documents and document sub-parts;
- W3C has defined accompanying standards for inter- and intra-document linkage (XPath, XPointer, and Xlink) as well as document traversal and transformation (XSLT);
- XML is fully integrated with emerging standards such as the Resource Definition Framework (RDF) and DAML+OIL, which can be "layered" on top of XML documents to provide a formal semantics defining XML-instantiated objects and relations.

We have defined an XML format for representing linguistic annotations called the *Generic Mapping Tool (GMT)*. The GMT defines XML elements for encoding annotation structure (primarily, a nestable `<struct>` element) and data categories (a nestable `<feat>` tag). A `<seg>` element provides a pointer to the annotated data using XPointers. Relations among objects can be specified explicitly using a `<rel>` element or may be implicit in the hierarchical nesting of `<struct>` elements. The GMT is described in detail in Ide & Romary, 2001b. We stress, however, that the details of the XML format—in particular, element names—is arbitrary; the only requirement is that the underlying data model can be expressed using the format.

### 5.1. Morpho-syntactic annotation

Morpho-syntactic annotation provides a good example of how the data model instantiated in the GMT is applied, and demonstrates some of the mechanisms required for representing annotations in general. Morpho-syntactic annotation involves the identification of word classes over a continuous stream of word tokens. The annotations may refer to the segmentation of the input stream into word tokens, but may also involve grouping together sequences of tokens or identifying sub-token units (or morphemes), depending on the language under consideration and, in particular, the definitions of "word" and "morpheme" as applied to this language. The description of word classes may include one or several features such as syntactic category, lemma, gender, number etc., which is again dependent on the language being analyzed.

Morpho-syntactic annotation can be represented by a single type of structural node (named W-level) representing a word-level structure unit. One or several information units are associated with each structural node.

For the purposes of illustration, we identify the following data categories (in practice these would be defined in reference to categories in the Data Category Registry):

/lemma/: contains or points to a reference word form for the token or sequence of tokens being described;
/part of speech/: a reference to a morpho-syntactic category;
/confidence/: a confidence level assigned by the manual or automatic annotator in ambiguous cases.
/gender/: the grammatical gender information associated with a word token or a sequence of word tokens;
/number/: the grammatical gender information associated with a word token or a sequence of word tokens;
/tense/: the grammatical tense information associated with a word token or a sequence of word tokens;
/person/: the grammatical person information associated with a word token or a sequence of word tokens.

The following provides an example of the morpho-syntactic annotation of the sentence "Paul aime les croissants" in the GMT format:[1]

```
<struct type="MSAnnot">
 <struct type="W-level">
   <feat type="lemma">Paul</feat>
   <feat type="pos">PNOUN</feat>
     <seg target="#w1"/>
 </struct>
 <struct type="W-level">
   <feat type="lemma">aimer</feat>
   <feat type="pos">VERB</feat>
   <feat type="tense">present</feat>
   <feat type="person">3</feat>
   <seg target="#w2"/>
 </struct>
```

---

[1] For brevity, we use an abbreviated pointer syntax to refer to the primary data in this example.

```
    <struct type="W-level">
      <feat type="lemma">le</feat>
      <feat type="pos">DET</feat>
      <feat type="number">plural</feat>
      <seg target="#w3"/>
    </struct>
    <struct type="W-level">
      <feat type="lemma">croissant</feat>
      <feat type="pos">NOUN</feat>
      <feat type="number">plural</feat>
      <seg target="#w4"/>
    </struct>
</struct>
```

Note that there is no limit to the number of information units that may be associated with a given structural node (as opposed to the text based representations that are usually provided by available POS taggers). It is also possible to structure the annotations by embedding `<feat>` elements to reflect a more complex feature-based annotation, or by pointing to a lexical entry providing the information.

In some cases, the morpho-syntactic annotation of a word or sequence of words requires a hierarchy of word level structures (e.g., when a word token results from the combination of several morphemes that must be annotated independently). For example, some occurrences of the token "du" in French can be analyzed as the fusion of the preposition "de" with the determiner "le" (as in "la queue *du* chat"). This is handled by embedding word-level structures as follows:

```
<struct type="W-level">
  <seg target="#w1"/>
  <struct type="W-level">
     <feat type="lemma">de</feat>
     <feat type="pos">PREP</feat>
  </struct>
  <struct type="W-level">
     <feat type="lemma">le</feat>
     <feat type="pos">DET</feat>
  </struct>
</struct>
```

Conversely, annotation of compound words may involve associating a single lemma to a sequence of word tokens at the surface level. In this case, the lemma is attached to the higher level of embedding and reference to the source is given at the leaves of the hierarchy, as in the following representation of the compound "pomme de terre" in French :

```
<struct type="W-level">
  <feat type="lemma">
       pomme_de_terre</feat>
  <feat type="pos">NOUN</feat>
  <struct type="W-level">
     <seg target="#w1"/>
     <feat type="lemma">pomme</feat>
     <feat type="pos">NOUN</feat>
  </struct>
  <struct type="W-level">
     <seg target="#w2"/>
     <feat type="lemma">de</feat>
     <feat type="pos">PREP</feat>
  </struct>
  <struct type="W-level">
     <seg target="#w3"/>
     <feat type="lemma">terre</feat>
     <feat type="pos">NOUN</feat>
```
```
  </struct>
</struct>
```

The ability to specify a hierarchical structure where needed enables specification of the level of granularity required. This is especially critical for a representation scheme, since the granularity of the segmentation in (or associated with) the primary data may not directly correspond to the level of granularity required for the annotation.

### 5.1.1. Alternatives

Morpho-syntactic annotation can be used to illustrate the representation of both structural and informational alternatives, which arises when a given word token is associated with two or more word classes. For example, the French word "bouche" which can be derived both from the verb "boucher" and the noun "bouche", which can be represented as follows:

```
<struct type="W-level">
  <seg target="#w1"/>
  <alt>
    <feat type="lemma">boucher</feat>
    <feat type="pos">VERB</feat>
    <feat type="tense">present</feat>
    <feat type="confidence">0.4</feat>
  </alt>
  <alt>
    <feat type="lemma">bouche</feat>
    <feat type="pos">NOUN</feat>
    <feat type="confidence">0.6</feat>
  </alt>
</struct>
```

### 5.1.2. Relating annotation levels

We assume the use of stand-off annotation; that is, an annotated corpus is represented as a lattice of stand-off annotation documents pointing to a primary source or intermediate annotation levels. However, depending on the point of view, the relations between various annotation levels can be more or less explicit. It is possible to identify three major ways to relate different levels of annotation: temporal anchoring, event-based anchoring, and object-based anchoring.

Temporal anchoring associates positional information to each structural level. This positional information is typically represented as a pair of numbers expressing the starting point and ending point of the segment being described. To do so in our framework, we introduce two attributes for the `<seg>` element:

/startPosition/: the temporal or offset position of the beginning of the current structural node;

/endPosition/: the temporal or offset position of the end of the current structural node.

For example, the following associates a phonetic transcription with a given portion of a primary text:

```
<struct type="phonetic">
  <seg startsAt="2300"
       endsAt="3200"/>
  <feat type="phone">iy</feat>
</struct>
```

We also define an event-based anchoring, which effectively introduces a structural node to represent a location in the text, to which all annotations for the object

at that location can refer. This strategy is useful in two cases:

> Situations where it is not possible or desirable to modify the primary data by inserting markup to identify specific objects or points in the data (e.g., speech annotation, associated with a speech signal, or in general any "read-only" data).
>
> Primary data marked with "milestones", such as time stamps in speech data, where spans across the various milestones must be identified. In this case, the `<struct>` elements represent the markup for segmentation (e.g., segmentation into words, sentences, etc.).[2]

To represent this, we introduce a specific type of structural node, named *landmark*, which is referred to by annotations for the defined span, as follows:

```
<struct type="landmark">
   <seg startsAt="2300"
        endsAt="3200"/>
</struct>
```

The third mechanism, object-based anchoring, enables pointing from a given level to one or several structural nodes at another level. This mechanism is particularly useful to make dependencies between two or more annotation levels explicit. For example, syntactic annotation can refer directly to the relevant nodes in a morpho-syntactically annotated corpus, in order, for example, to identify the correct NP "le chat" in "la queue du chat", as shown below:

```
<!-- Morphosyntactic level -->
<struct type="W-level">
   <seg target="#w3">
   <struct type="W-level">
     <seg target="#w3.1">
     <feat type="lemma">de</feat>
     <feat type="pos">PREP</feat>
   </struct>
    <struct type="W-level">
      <seg target="#w3.2">
      <feat type="lemma">le</feat>
      <feat type="pos">DET</feat>
      <feat type="gender">masc</feat>
   </struct>
  </struct>
   <struct type="W-level">
      <seg target="#w4">
      <feat type="lemma>chat</feat>
      <feat type="pos">NOUN</feat>
   </struct>
</struct>
<!-- Syntactic level (simplified) -->
<struct>
   <feat type="synCat">NP</feat>
   <seg targets="w3.2 w4"/>
</struct>
```

---

[2] The annotation graph (AG) formalism (Bird and Liberman, 2001) was explicitly designed to deal with time-stamped data. However, we feel the AG is not sufficiently general because (1) AG reifies the "arc" and distinguishes it from identification of spans via, e.g., XML tags; and (2) AG requires *ad hoc* mechanisms to deal with hierarchically organized annotations. In both cases, AG requires different mechanisms to treat analogous constructs.

## 5.2. Summary

The framework presented here for linguistic annotation is intended to allow for variation in annotation schemes while at the same time enabling comparison and evaluation, merging of different annotations, and development of common tools for creating and using annotated data. We have developed an abstract model for annotations that is capable of representing the necessary information while providing a common encoding format that can be used as a pivot for combining and comparing annotations, as well as an underlying format that can be manipulated and accessed with common tools. The details presented here provide a look "under the hood" in order to show the flexibility and representational power of the abstract scheme. However, the intention is that annotators and users of annotation schemes can continue to use their own or other formats with which they are comfortable; as long as the underlying data model is the same, translation into and out of this or any other instantiation of the abstract format will be automatic.

Our framework for linguistic annotation is built around some relatively straightforward ideas: separation of information conveyed by means of structure and information conveyed directly by specification of content categories; development of an abstract format that puts a layer of abstraction between site-specific annotation schemes and standard specifications; and creation of a Data Category Registry to provide a reference set of annotation categories. The emergence of XML and related standards, such as RDF, provides the enabling technology. We are, therefore, at a point where the creation and use of annotated data and concerns about the way it is represented can be treated separately—that is, researchers can focus on the question of *what* to encode, independent of the question of *how* to encode it. The end result should be greater coherence, consistency, and ease of use and access for linguistically annotated data.

## 6. Conclusion

ISO TC37/SC4 is just beginning its work, and will use the general framework discussed in the preceding sections as its starting point. However, the work of the committee will not be successful unless it is accepted by the language processing community. To ensure widespread acceptance, it is critical to involve as many representatives of the community in the development of the standards as possible, in order to ensure that all needs are addressed. This paper serves as a call for participation to the language processing community; those interested should contact the TC 37/SC 4 chairman (Laurent Romary: romary@loria.fr).